\documentclass{article}

\usepackage[preprint]{nips_2018}

\usepackage[utf8]{inputenc} % allow utf-8 input
\usepackage[T1]{fontenc}    % use 8-bit T1 fonts
\usepackage{hyperref}       % hyperlinks
\usepackage{url}            % simple URL typesetting
\usepackage{booktabs}       % professional-quality tables
\usepackage{amsfonts}       % blackboard math symbols
\usepackage{nicefrac}       % compact symbols for 1/2, etc.
\usepackage{microtype}      % microtypography

\usepackage{amsmath,amsfonts,amssymb}
\usepackage{graphicx}

\title{ExIt-OOS: Towards Learning from Planning in Imperfect Information Games}

\author{
  Andy C.~Kitchen \\
  Head of AI Research \\
  CliniCloud Labs \\
  Melbourne, Australia VIC 3000 \\
  \texttt{andy.kitchen@clinicloud.com} \\
  \And
  Michela Benedetti \\
  Machine Learning Engineer \\
  CliniCloud Labs \\
  Melbourne, Australia VIC 3000 \\
  \texttt{michela.benedetti@clinicloud.com} \\
}

\begin{document}

\maketitle

\begin{abstract}
  The current state of the art in playing many important perfect
  information games, including Chess and Go, combines planning and
  deep reinforcement learning with self-play. We extend this approach
  to imperfect information games and present ExIt-OOS, a novel
  approach to playing imperfect information games within the Expert
  Iteration framework and inspired by AlphaZero. We use Online Outcome
  Sampling, an online search algorithm for imperfect information games
  in place of MCTS. While training online, our neural strategy is used
  to improve the accuracy of playouts in OOS, allowing a learning and
  planning feedback loop for imperfect information games.
\end{abstract}

\section{Introduction}

Many recent gains in game playing skill for perfect information games
have come from combining planning, deep reinforcement learning and
self play. In the Expert Iteration and AlphaZero framework, a powerful online
planning/search algorithm, usually Monte Carlo Tree Search (MCTS)
\citep{browne2012survey} is combined with a learnable heuristic value
and policy function, represented with a deep neural network. During
self-play, the learned heuristics are used to guide the planner. The
action recommended by the time consuming planning process is used as
feedback to train the heuristic function. As this heuristic improves,
so does the quality of actions chosen by the planner. When executed
carefully this leads to a cycle of mutual improvement and very strong
play. Notably these approaches require no expert domain knowledge of
the game to be manually incorporated. The recent success of the
AlphaZero \citep{silver2017mastering} and Expert Iteration (ExIt)
\citep{anthony2017thinking} exemplifies this approach.

The success of these methods can be partially explained by the
complementary nature of MCTS and Deep Neural Networks. DNNs and
Convolutional Neural Networks are powerful pattern recognisers which
can learn strategies that generalise well between states. However they
cannot roll out the combinatorial consequences of hypothetical
decisions. MCTS can bring consistency between state values and policy
choices by smoothly propagating the consequences of hypothetical
future decisions backwards up the game tree to previous nodes.

We present ExIt-OOS, an instance of Expert Iteration using Online
Outcome Sampling \citep{lisy2015oos}, an online planning
algorithm for imperfect information games, in place of MCTS. Thus
extending the learning-from-planning paradigm to imperfect information
games. This allows the playing of a wide class of imperfect information
games without modification, while making use of the rich information
provided by planning. During search the current neural strategy is
used during the rollout phase of OOS to improve the quality of the
search. We present experimental data on exploitability and head to
head matches between OOS and neutral nets trained with ExIt-OOS.

\section{Background}

\subsection{Imperfect Information in Extensive-Form Games}

Extensive-form games are annotated trees that model sequential
decision making with multiple players. Each node in the tree
represents a game state and is labelled with a player. The edges
leading out from each node represent actions that can be taken by the
acting player in that state. Aside from the active players, there is
also an auxiliary chance player who always takes actions with a fixed
probability, modelling randomness in the game.  The leaf nodes
represent terminal states and carry a vector of utilities, one for
each player. Players conventionally choose their strategies to
maximise utility. In a zero-sum game, the utilities at each terminal
node sum to zero.

In an imperfect information game, many states are indistinguishable to
a given player and are grouped into information sets (infosets). We
only consider games with perfect recall. Where players always remember
all information that has been revealed to them and every action they
have taken in the past. States with different observable histories
must be in different infosets. A behaviour strategy is a mapping from
infosets to probabilities over actions. A player must behave in the
same way for every state in an infoset. Games are provided to our
ExIt-OOS implementation as programs that implicitly define the
extensive form game tree.

\subsection{Nash Equilibrium}

When playing imperfect information games some care is required in the
definition of an optimal strategy as discussed in
\cite{billings2004game}, from a game theoretic perspective, optimal
usually means playing a Nash equilibrium strategy, where no player has
any incentive to unilaterally change their strategy, given the
strategy of the other players. However this can be problematic,
because this approach can be overly defensive; the opponent is often
fallible and has flaws that can be exploited. For example, in a poker
tournament, players with strategies far from equilibrium --- but able
to exploit weaker entrants --- may well finish ahead of equilibrium
players (in another sense, the `meta-game' of changing strategies
between hands against realistic opponents is not modelled
correctly). However, this work focuses on finding equilibrium strategies
for single game instances and leaves online adaptation, opponent
modelling and exploitation to future work.

\subsection{Online Outcome Sampling and Targeting}

Online Outcome Sampling is the first published imperfect information
search algorithm that converges to equilibrium strategies in
two-player zero-sum games \citep{lisy2015oos}. It is a sampling
algorithm that uses regret matching to minimise the counterfactual
regret locally at each infoset in a game tree. It is essentially MCCFR
with incremental tree growth and targeting.

In the Online Outcome Sampling algorithm, all simulations are run from
the initial state to a terminal state, simulations are dynamically
targeted to infosets observed during play with boosted
probability. When a simulation is targeted, importance sampling is
used to reweight targeted simulations to keep collected statistics
unbiased. The OOS authors propose two kinds of targeting. Firstly, Public
Set Targeting (PST) where simulations are forced to be consistent with
the public subgame so far i.e. players always take their observed
public actions and public chance nodes take their observed
outcomes. But private information such as the current hand in poker
may differ.  Secondly, Information Set Targeting (IST) where simulation
actions are forced to be consistent with all current observations of
the game in progress, including private information.  Implementing
targeting may require providing additional complex domain knowledge,
beyond just the dynamics of the game to be played. Observed
information may constrain the actions an opponent could have taken in
complex ways, for example, targeting in II Goofspiel requires
constraint solving.

Strong play in imperfect information games presents challenges not
present when playing perfect information games. Firstly, concealing
private information, playing unpredictability, and even misdirection
and deception become important factors. If a player is too predictable
the opponent may exploit this weakness. Furthermore, the value at a
given infoset, depends on the distribution over states conditioned on
reaching that infoset, which in turn is dependent on the strategy both
players have been using since the initial state. In imperfect
information games counterfactuals matter. Past games that did not
happen, but could have happened affect current strategic choices. OOS
and other counterfactual regret minimisation algorithms employ careful
weighting to ensure these factors are taken into account. Changes in
strategy in one place in the game tree, can cause the best response in
another far away area to change, this is called non-locality
\citep{lisy2015oos}. OOS keeps statistics and continually
updates strategies even for infosets that are no longer reachable.

\subsection{Expert Iteration}

Many powerful search and planning algorithms, such as MCTS are table
driven and collect statistics per state or infoset during
planning. They cannot generalise and share information between states
even when they are functionally very similar in the
game. Contrastingly, Deep Neural Networks can learn function
approximations that generalise well, but creating high quality
training data is difficult, if the correct strategic choice could
already be calculated precisely this procedure could be used to
play the game well and the problem would be solved. If computation
speed was the only concern, Imitation Learning could be used to train
a neural network to imitate the results of a much slower search, but
it could not surpass the quality of the base search that it was
learning from. The key idea of Expert Iteration (ExIt)
\citep{anthony2017thinking} is that the search procedure can be
improved with feedback from a partially trained apprentice neural
network, creating a cycle of mutual improvement. A better heuristic,
creates more accurate training targets which further improves the
heuristic.

\subsection{AlphaZero}

The AlphaZero \citep{silver2017mastering} algorithm trains a deep
neural network to approximate a value function and policy using
targets generated from MCTS and self-play. Demonstrating
state-of-the-art play in Chess, Go and Shogi with no human provided
heuristic functions. Although AlphaZero did require extensive
meta-domain knowledge in the choice of neural network architecture and
hyper-parameter selection. AlphaZero uses a sophisticated CNN
architecture to encode board positions and move selection, also
bringing some significant structural prior information.

AlphaZero learns to approximate a value and policy function with a
deep neural network:

$$ (\mathbf{p}, v) = f_{\theta}(s) $$

Where $s$ is an encoding of the current state, $v$ is a scalar state
value, $p$ is a probability vector over actions and $\theta$ is a
vector of neural network parameters. The policy network is trained on
a distribution proportional to the visit counts at the root of MCTS
searches, the value network is trained with the utility of the
completed game. AlphaZero uses the neural approximation to improve the
MCTS search in two ways. First, during the expansion step, the neural
network is evaluated and used to compute prior action probabilities
for new nodes. Second, instead of a rollout phase, the neural network
value function at the tree fringe is used as a surrogate and
backpropagated up the tree. Another perspective is that the Monte
Carlo search is a policy improvement operator applied to the current
neural policy. It is evaluated at states sampled from self play, while
the value network takes the role of policy evaluation making
AlphaZero a unconventional and elaborate, policy iteration algorithm.

\section{ExIt-OOS}

ExIt-OOS is instance of Expert Iteration where Online Outcome Sampling
is used for planning and takes the role of the expert, The apprentice
neural network is used as a playout policy to improve the decisions of
the expert. The use of Online Outcome Sampling allows planning online
in imperfect information games, and can converge to a Nash
equilibrium.

One major difference between the operation of MCTS and OOS is that in
the latter, node strategies are constantly changing and never converge
to a specific value, only the average strategy at the node
converges. Unfortunately this protean aspect prevents the simple
introduction of node priors, although work in this area is ongoing. In
our implementation the neural network policy is used to provide
accurate rollouts. We learn a neural strategy:

$$ \mathbf{p} = f_\theta(I_s)
\qquad l = \sigma^\intercal(\log \sigma - \log \mathbf{p}) $$

Where $\mathbf{p}_a$ is the probability of taking action $a$ given the
infoset encoding $I_s$ at state $s$, $f_\theta(I_s)$ is a deep neural
network with parameters $\theta$. The loss $l$ is a KL Divergence
between the OOS expert provided target and the neural neural network
output.

Our high performance implementation runs many concurrent simulations
which are represented by state machines. Each simulation's state
machine is progressed until it enters a state that requires neural net
evaluation, at which point it is added to a evaluation waiting
list. When there a no simulations that can progress any further, the
waiting list is turned into a batch which is evaluated by the neural
network. All simulations can then continue to progress until the next
evaluation cycle.

Our experience generation is also run in parallel, where one process
is launched for every core. Each process plays one game from start to
finish and collects experience and planning results in a list of
experience tuples:

$$ (I_s, \sigma_\text{oos})_i \qquad i = 1, \ldots, N $$

Which are composed of the infoset encoding $I_s$ and the action
probabilities $\sigma_\text{oos}$ found by OOS. $N$ is the number of
experience tuples generated. These are collected in a training process
which adds these tuples to the experience reservoir and does a number
of steps with gradient descent before sending the new updated neural
network parameters to each simulation processes. We use an
exponentially weighted reservoir of experience similar to
\cite{heinrich2016deep} where new incoming experiences replace old
experiences with a certain fixed probability. After each batch of
experience from the game playing processes comes in, minibatches are
sampled from the reservoir for training.

\section{Experiments}

We tested our algorithm on Leduc poker \citep{southey2012bayes}, II
Goofspiel with 6 cards and 13 cards and Liar's Dice with 1 or 2 dice
\citep{lisy2015oos}. Leduc poker is a highly simplified research poker
variant.  Leduc poker has 6 cards, two suits and three ranks King,
Queen, Jack. Each player has a one card hand and there is one
community card. The are two betting rounds one before and after the
community card is revealed. There is an ante of 1, the first round has
a fixed bet/raise of 2 followed in the second by 4. There is a maximum
of 2 raises per round. There are only two hand types: high card, or
pair formed with the community card.  Draws are worth 0, otherwise the
utility is the bet value gained or lost after the showdown or when a
player folds. All the infosets in Leduc poker are the same size and
have 3 elements. Leduc poker has 288 distinct infosets. We use Public
Set Targeting in experiments with Leduc Poker.

II Goofspiel($N$) is a variant of Goofspiel. Each player starts with a
hand of cards with rank $0, 1, \ldots N-1$. There are $N$ rounds, at
each round a card with point value $P$ is revealed and both players
bid simultaneously on that card by choosing a card from their
hand. The player with the higher bid is awarded $P$ points and both
bid cards are removed from their players hands. If both bids are the
same rank, no points are awarded. In II Goofspiel, the value cards are
revealed in a fixed increasing order, $P = 0, 1, \ldots N-1$ and only
the winners of each bidding round are revealed, not the rank of the
bid.  Infosets in II Goofspiel have differing sizes. The infosets
increase rapidly in size as the game goes on, while tapering closer to
the end, they contain every possible combination of remaining opponent
cards, given the rank constraints observed on previous rounds. II
Goofspiel($6$) has order $10^5$ infosets and II Goofspiel($13$) has
approximately $10^9$ infosets. Information Set Targeting is used for
Goofspiel experiments, there is no public game tree.

Liar's Dice($N$) is a bluffing game played with $N$, 6-sided dice for
each player. 1-5 are normal face values, 6 is wild. Each player throws
their dice and keeps the outcome private. On each players turn they
may either bid, or call ``liar!''. Each bid is a quantity and face
value, e.g. ``a pair of fours'' or ``three dice showing 6''. Bids must
always be greater in quantity or value than the opponent's last
bid. When liar is called, if the last bid doesn't hold (with the wild
6 matching any number) the caller wins; otherwise if the bid holds,
the caller loses.

We use hyper parameters from \cite{lisy2015oos}, in all
experiments, The search exploration parameter $\epsilon = 0.4$, the
targeting probability $\delta = 0.9$, and the opponent mistake
probability $\gamma = 0.01$. The only change we make is that instead
of manually reweighting our samples when they are targeted, an
exponential moving average with decay $\beta = 0.99$ is used to
estimate $ r = \mathbb{E}[s_1 / s_2] $ where $r$ is the ratio of $s_1$
the probability that a state is sampled with targeting and $s_2$ the
probability that a state is sampled without targeting with expectation
taken over all simulations. Multiplying importance weights by $1/r$
puts heavier weight on samples from later in the game where there is
heavier targeting. Our Leduc poker neural network has a single hidden
layer with 128 hidden neurons.  Our II Goofspiel(6) net 2 had hidden
layers with 128 and 64 neurons, in II Goofspiel(13) experiments we
used 3 hidden layers of 128, 128 and 64. ReLu activation functions are
used for all nets. We use a softmax output layer large enough to
contain all legal actions. Illegal actions have their probabilities set
to zero while the rest are normalised, they are not considered in the
loss calculation. We use a reservoir that can hold experience from
$32{,}000$ full games with a decay ratio of $\beta_\text{reservoir} =
2$. In Leduc and Goof'(6) we train 128 steps every 32 games, for
Goof'(13) we train 512 steps every 32 games. All head-to-head match ups
were run for 5000 games. One iteration corresponds to 32 episodes, and
128 or 512 gradient descent steps, we use Adam as our optimisation
algorithm and a learning rate of $10^{-3}$. We used 20 concurrent
searches in all experiments. In the Goof'(6) and Goof'(13) experiments,
we ran 10k simulation steps.

\begin{figure}
  \begin{center}
    \include{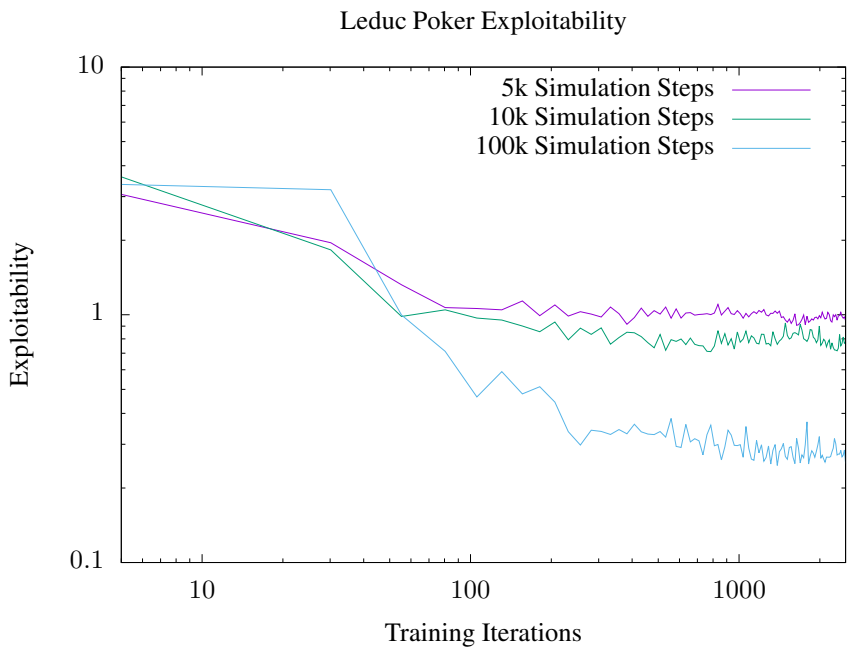}
  \end{center}
  \caption{Leduc Poker Exploitability During Training}
  \label{fig:leduc-ex}
\end{figure}

In figure \ref{fig:leduc-ex} we see the exploitability during training
in Leduc poker, performance is very dependent on how many search
simulations are run, however Leduc poker has such a small number of
infosets that the whole tree is created very quickly, so the prior
knowledge from rollouts doesn't help much.  In table
\ref{table:goof6win} with Goof'(6) we again see performance reaching a
ceiling in head to head match ups with the teacher. For Goof'(13) and
Liar's Dice(2) in tables \ref{table:goof13win} and \ref{table:liars2win},
we see increasing performance, eventually slightly surpassing the
teacher; we believe this is due to the much greater value provided by
rollouts in the larger game. All neural networks are feed forward and
execute orders of magnitude faster than running the search.

\begin{table}[]
  \vspace{1.5cm}
  \begin{center}
    \begin{tabular}{| r r | l l |}
\hline
II Goof'(6) & Iter. & Random & OOS (10k) \\
\hline
ExIt-OOS  &  100 & 66\% (2.5) & 30\% (2.7) \\
          &  200 & 63\% (2.6) & 34\% (2.7) \\
          & 1000 & 69\% (2.5) & 26\% (2.2) \\
          & 2000 & 68\% (2.5) & 27\% (2.2) \\
          & 2500 & 67\% (2.5) & 25\% (2.2) \\
\hline
    \end{tabular}
  \end{center}
  \vspace{0.5cm}
  \caption{Win rates of NN trained with
    ExIt-OOS for II Goofspiel(6)}
  \label{table:goof6win}
\end{table}

\begin{table}[]
  \begin{center}
    \begin{tabular}{| r r | l l |} 
\hline
II Goof'(13) & Iter. & Random & OOS (10k) \\
\hline
ExIt-OOS  & 100 & 61\% (2.7) & 40\% (2.7) \\
          & 200 & 69\% (2.5) & 44\% (2.7) \\
          & 500 & 74\% (2.4) & \bf{53\%} (2.6) \\
\hline
    \end{tabular}
  \end{center}
  \vspace{0.5cm}
  \caption{Win rates of NN trained with
    ExIt-OOS for II Goofspiel(13)}
\label{table:goof13win}
\end{table}

\begin{table}[]
  \begin{center}
    \begin{tabular}{| r r | l l |} 
\hline
Liar's Dice(1) & Iter. & Random & OOS (10k) \\
\hline
ExIt-OOS  & 100 & 65\% (2.6) & 53\% (2.8) \\
          & 500 & 65\% (2.6) & 53\% (2.8) \\
          & 1000 & 66\% (2.6) & 51\% (2.8) \\
\hline
    \end{tabular}
  \end{center}
  \vspace{0.5cm}
  \caption{Win rates of NN trained with
    ExIt-OOS for Liar's Dice(1)}
\label{table:liars1win}
\end{table}

\begin{table}[]
  \begin{center}
    \begin{tabular}{| r r | l l |} 
\hline
Liar's Dice(2) & Iter. & Random & OOS (10k) \\
\hline
ExIt-OOS  & 100 & 65\% (2.6) & 52\% (2.8) \\
          & 500 & 64\% (2.7) & 52\% (2.8) \\
          & 1000 & 66\% (2.6) & 54\% (2.8) \\
\hline
    \end{tabular}
  \end{center}
  \vspace{0.5cm}
  \caption{Win rates of NN trained with
    ExIt-OOS for Liar's Dice(2)}
\label{table:liars2win}
\end{table}

\begin{table}[]
  \begin{center}
    \begin{tabular}{| r r | l |} 
\hline
Alg. & Iter. & II Goof'(13) Elo \\
\hline
ExIt-OOS  & 100 & 1500 (\emph{fixed}) \\
          & 200 & 1531 (4.1) \\
          & 500 & \bf{1600} (3.5) \\
\hline
OOS 10k   & & 1572 (3.7) \\
\hline
    \end{tabular}
  \end{center}
  \vspace{0.5cm}
  \caption{Elo rating of NN trained with
    ExIt-OOS for II Goofspiel(13)}
  \label{table:goof13elo}
\end{table}

\section{Related Work}

Expert Iteration (ExIt) was independently developed from the similar
AlphaZero. The algorithm was used to improve the state-of-the-art in
game Hex \citep{anthony2017thinking}. The authors of the algorithm
propose an analogy to human psychology and ``thinking fast and slow''
where system 1 and system 2 thinking integrate to solve problems. Our
algorithm is heavily inspired by AlphaZero and is an instance of ExIt,
so therefore is related to AlphaGo and AlphaGo Zero. These algorithms
all combine planning, self-play and deep learning. However AlphaGo
also used supervised learning from human games and did not use the
outcome of MCTS planning as training targets. While very general,
without some modification, none of these algorithms are suitable for
playing imperfect information games.

The state of the art for playing imperfect information games, are
exemplified by the AI poker playing programs Libratus
\citep{brown2017libratus} and DeepStack
\citep{moravvcik2017deepstack}, both use safe subgame re-solving as a
core component. Libratus, uses a hybrid offline and online
method. Where a relatively coarse abstract game is solved offline with
Monte Carlo Counter Factual Regret minimisation to create a base
strategy. The base strategy is refined online using safe subgame
re-solving with a fine grained abstraction. The subgame fixes
all betting actions and community cards so far, with modified hand
probabilities reflecting the relative chance of having a certain hand
conditioned on reaching this subgame. DeepStack never computes and
stores a full strategy and instead uses continual re-solving for every
decision.  Deep counterfactual value networks are used to approximate
search results after a certain search depth limit. Their neural
network was trained supervised on a dataset of randomly generated
games that were solved offline. DeepStack and Libratus are built from
general building blocks, but rely on the availability of a rich
subgame structure.  This is present in poker because all actions are
public, forming a very informative public game tree. They also gain
efficiency from the fact that all Heads-Up Texas Hold'em Poker
infosets are the same size, have the same structure, and are
relatively small (the only unknown information is a a combination of 2
opponent playing cards). DeepStack does not have a feedback loop
between planning and training, the deep counterfactual value network
is trained on fixed offline solutions. Libratus is table driven and
does not utilise neural networks for generalisation.

Our approach is similar to Neural Fictitious Self Play (NFSP)
\citep{heinrich2016deep} in the use of deep neural networks to
represent policies and training on experiences generated during self
play. NFSP uses a Deep Q-learning like algorithm to approximate a best
response in a model-free way. It does not use any planning. The
benefit of using Q-learning is that the NFSP does not require access
to the dynamics of the environment. However if model dynamics are
available, NFSP is unable to benefit. We do not use fictitious self
play, and only keep one current best estimate of the Nash equilibrium
strategy learned from the online search.

SmoothUCT \citep{heinrich2015smooth}, and MCTS with Exp3 or Regret
Matching for action selection \citep{lisy2013convergence} are planning
methods for imperfect information games that can converge to Nash
equilibria in large complex games, but must be run in an offline
setting with each simulation run from the root.  Online variants of
these algorithms may later be created and could be incorporated into
ExIt-OOS. Information Set Monte Carlo Tree Search (ISMCTS)
\citep{cowling2012ismcts} is an extension of MCTS which can be run
online in imperfect information settings, however ISMCTS samples
hidden states uniformly from the current infoset, and may produce
highly exploitable strategies, however ISMCTS works well in practice
for many games.  All these planning methods keep statistics per
infoset and have no generalisation between states. Hand crafted
abstractions are often used when solving large games to reduce the
search space by aliasing similar states and actions together, thus
manually sharing information between states and actions --- instead of
learning from data.

\section{Future Work}

Online Outcome Sampling is an effective online search algorithm.
However even with high quality rollouts provided by the companion deep
neural network, the exploitability in seems to be somewhat limited by
the number of search iterations. An intriguing alternative is to use
fictitious self play in a similar vein to NFSP, but replace the Neural
Q-learning learning with a best response computed using a one sided
MCTS search (where the other player strategy is kept fixed at the
average strategy). Although, even with the opponent strategy
fixed computing a best response still requires solving a POMDP
online. It is possible the distribution over opponent hidden states
could be learned and sampled for each search iteration.

There is also the possibility that due to the highly non-local nature
of equilibrium play; small changes in strategy in one place in the
game tree can cause large changes at distant nodes in response. Local
search systems may have their limits. Global black box,
optimisation algorithms like evolutionary strategies may be more
suitable and general. Perhaps these algorithms could be adapted to
converge to a robust and balanced population of strategies. This could
lead to better play even for games without a well defined Nash
equilibria such as full table poker. Having a balanced population of
agents with interlocking strengths and weaknesses, chosen in the
correct way could allow a meta-game playing strategy to choose the
best one to exploit opponents over repeated rounds.

Ultimately there may be some combination of techniques similar to
ExIt-OOS and AlphaZero that allow building an AI computer program
that can reach super-human performance on any given abstract strategy
game even with imperfect information. Using a single learning algorithm.
Without any human domain knowledge. This is a small step in that
direction. The authors are excited to continue this work.

\section{Conclusion}

The ExIt-OOS algorithm presents a novel approach to learn neural
strategies in a large and general class of imperfect information
games, even when there is no subgame structure. ExIt-OOS can utilise
the high quality strategic information derived from planning with
Online Outcome Sampling and known environment dynamics.

\section{Acknowledgements}

The authors would like to thank anonymous colleagues for implementing
a large part of our testing infrastructure. GNU parallel was used
for some experimental runs \citep{tange2018}. PyTorch was
a major implementation component \citep{paszke2017automatic}.

\bibliography{oz}

\end{document}